\documentclass[letterpaper]{article} 
\usepackage{aaai23}  
\usepackage{times}  
\usepackage{helvet}  
\usepackage{courier}  
\usepackage[hyphens]{url}  
\usepackage{graphicx} 
\usepackage{natbib}  
\usepackage{caption} 
\usepackage{algorithm}
\usepackage{algorithmic}
\usepackage{booktabs}  
\usepackage{newfloat}
\usepackage{listings}
\DeclareCaptionStyle{ruled}{labelfont=normalfont,labelsep=colon,strut=off} 
\floatstyle{ruled}
\newfloat{listing}{tb}{lst}{}
\floatname{listing}{Listing}
\title{Question Decomposition Tree for Answering Complex Questions over Knowledge Bases}
\author{
    Xiang Huang, Sitao Cheng, Yiheng Shu, Yuheng Bao, Yuzhong Qu
}
\affiliations{
    State Key Laboratory for Novel Software Technology, Nanjing University, China\\

    \{xianghuang, stcheng, yhshu, yhbao\}@smail.nju.edu.cn, yzqu@nju.edu.cn
    
}

\usepackage{bibentry}

\begin{document}
\newcommand{\tabincell}[2]{\begin{tabular}{@{}#1@{}}#2\end{tabular}}
\newcommand{\qdtrees}{\texttt{QDTrees}}
\newcommand{\cluedecipher}{\texttt{Clue-Decipher}}
\newcommand{\qdtqa}{\texttt{QDTQA}}
\newcommand{\qdtqasqg}{\texttt{QDTQA-SQG}}
\newcommand{\qdtqaseq}{\texttt{QDTQA-S2S}}
\newcommand{\lcqdtrees}{\texttt{QDTrees-LC}}
\newcommand{\cwqqdtrees}{\texttt{QDTrees-CWQ}}
\newcommand{\sota}{state-of-the-art}
\newcommand{\inq}{\texttt{[INQ]}}
\newcommand{\inql}{$[INQL]$}
\newcommand{\inqr}{$[INQR]$}
\newcommand{\des}{$[DES]$} 
\newcommand{\cluenet}{ClueNet} 
\newcommand{\deciphernet}{DecipherNet} 

\maketitle
\begin{abstract}

\begin{sloppypar}
Knowledge base question answering (KBQA) has attracted a lot of interest in recent years, especially for complex questions which require multiple facts to answer.
Question decomposition is a promising way to answer complex questions. 
Existing decomposition methods split the question into sub-questions according to a single compositionality type, which is not sufficient for questions involving multiple compositionality types.
In this paper, we propose Question Decomposition Tree (QDT) to represent the structure of complex questions.
Inspired by recent advances in natural language generation (NLG), 
we present a two-staged method called \cluedecipher~to generate QDT.
It can leverage the strong ability of NLG model and simultaneously preserve the original questions.
To verify that QDT can enhance KBQA task, we design a decomposition-based KBQA system called \qdtqa.
Extensive experiments show that \qdtqa~outperforms previous \sota~methods on ComplexWebQuestions dataset. 
Besides, our decomposition method improves an existing KBQA system by 11\% and sets a new \sota~on LC-QuAD 1.0.
\end{sloppypar}

\end{abstract}

\section{Introduction}
Question answering (QA) is a long-standing challenge in artificial intelligence. 
With the emergence of knowledge bases (KBs), such as DBpedia \cite{auer2007dbpedia} and Freebase \cite{bollacker2008freebase}, knowledge base question answering (KBQA) has attracted intensive attention \cite{lan2021survey}. 
However, answering complex questions with multiple hops or constraints is still challenging.
The difficulty of linking and compositing KB items (entities, relations, and constraints) grows intractably as questions become complex.  

\begin{figure}[tb]
    \centering
    \includegraphics[scale=0.67]{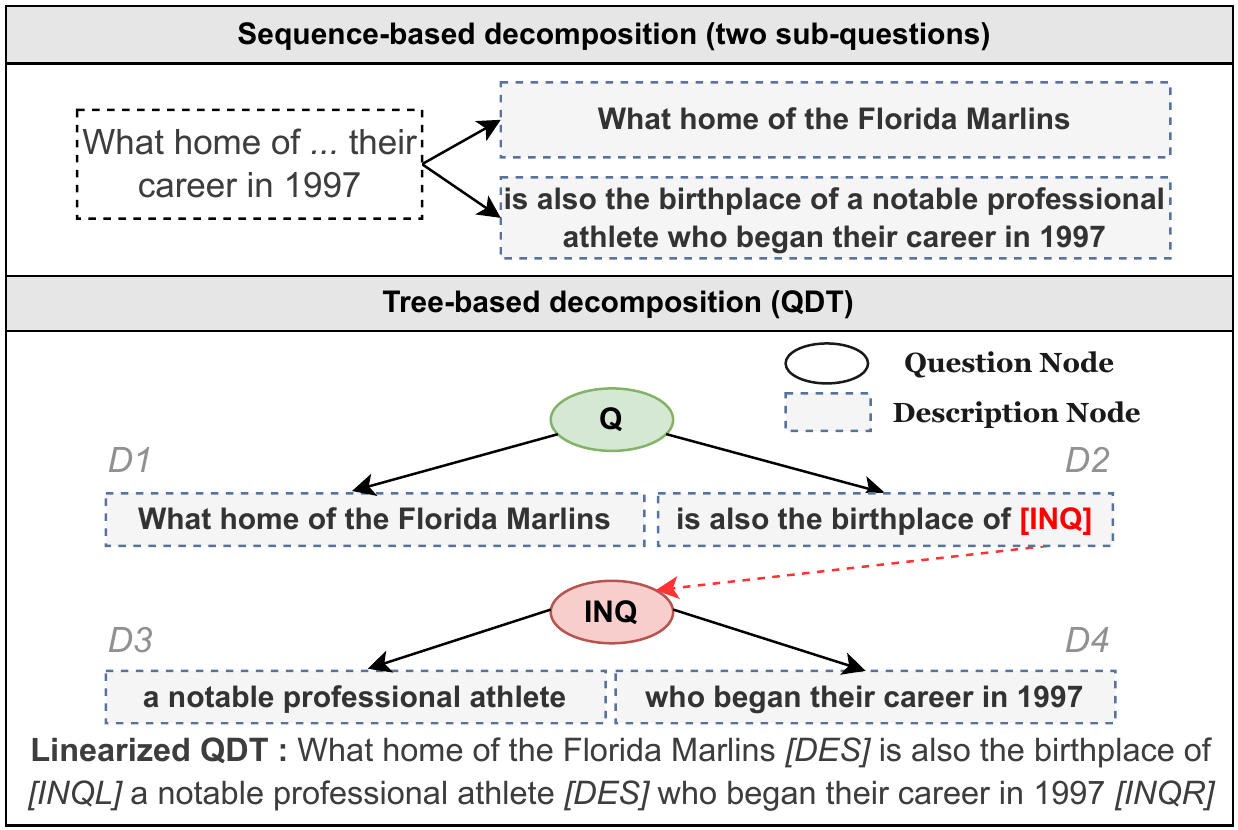}
    \caption{An example of Question Decomposition Tree (QDT) and sequence-based decomposition for the question “What home of the Florida Marlins is also the birthplace of a notable professional athlete who began their career in 1997?”. 
    [INQ] is a placeholder for an inner question.
    }
    \label{fig:main_example}
\end{figure}

To tackle the problems brought by complex questions, recent works \cite{Min2019multi,Talmor2018the,Zhang2019complex} put efforts into question decomposition and achieve remarkable performance. 
In these methods (sequence-based decomposition), a question is firstly classified into a single compositionality type, i.e., composition (with an inner question) or conjunction (with conjunctive descriptions).
Then, the question is split into two sub-questions according to its compositionality type.
For the exemplar question in Figure \ref{fig:main_example}, sequence-based methods would classify the question as a conjunction question and split it into two sub-questions (shown in the upper part).   
However, we find that 39\% and 31\% of the questions in ComplexWebQuestions 
 (CWQ) \cite{Talmor2018the} and LC-QuAD 1.0 (LC) \cite{trivedi2017lc}, contain more than one compositionality type, respectively.  
For the above example, the latter sub-question still contains an inner question, namely \textit{``a notable professional athlete who began their career in 1997''}, but sequence-based methods do not further decompose it, preventing this question from being answered correctly. 
EDGQA \cite{hu2021edg} alleviates such problem by decomposing the question into an entity-centric graph through hand-crafted rules.
But such graph structure is complicated and difficult to train with a neural model.

\begin{sloppypar}

In this paper, we focus on how to make question decomposition an effective way to answer complex questions over KBs.
We propose \textbf{Question Decomposition Tree (QDT)}, a tree-based decomposition structure to better model the structure of complex questions.
An example of QDT is shown in the lower part of Figure \ref{fig:main_example}. 
The root (green oval) is a question node that represents the whole question. 
It points to two description nodes, \textit{D1} and \textit{D2} (dashed rectangles).
\textit{D2} contains a placeholder ``[INQ]'' which indicates a sub-question.
In this manner, a QDT can represent the decomposition structure of a question with a combination of multiple compositionality types. 
We also construct a decomposition benchmark named \qdtrees~with 6,607 QDTs.

In order to generate QDT, we build our decomposition model \cluedecipher~based on NLG model and propose linearized QDT as target format, which makes it easier to generate a tree-based structure in a neural way.
Nevertheless, sometimes the generative method suffers from missing words or generating unexpected words (as shown in Table \ref{tab:seq2seqqd_error_case}).
\cluedecipher~addresses this issue by first generating a preliminary decomposition as a clue then inserting separators to the original question according to the clue.
It leverages the strong generation ability 
of NLG models and simultaneously ensures that the question remains unchanged.

Moreover, we design a QDT-based KBQA system called \qdtqa~to evaluate the effectiveness of QDT in answering complex questions over KB.
Experimental results show that \qdtqa~achieves \sota~result on CWQ.
In addition, our decomposition method improves an existing QA system by 11\% and sets a new \sota~on LC\footnote{Our code and dataset are available at \url{https://github.com/cdhx/QDTQA}}. 

\end{sloppypar}

\section{Related Work}
\label{sec:related work}

\subsection{Complex KBQA}
Complex KBQA aims at answering a question involving more than a single fact over KBs.
There are two mainstream methods to solve complex KBQA: information retrieval-based (IR-based) and semantic parsing-based (SP-based) methods \cite{lan2021survey}.
IR-based methods extract a question-specific subgraph from KB and then rank the candidate entities in the subgraph to get the final answer \cite{sun2019pullnet}.

SP-based methods fall into another line and generally perform better. 
They parse questions into formal queries that can be executed against KB.
Traditional SP-based methods follow a staged query graph generation manner, enumerating all possible query structures starting from a topic entity in limited hops, which introduces a large number of noisy candidates \cite{Yih2015Semantic,Chen2020Formal}.
Researchers have made efforts in pruning the search space.
QGG \cite{lan2020query} considers constraints in query generation and uses beam search to control the search space.
AQG \cite{Chen2020Formal} predicts an abstract query graph to restrict candidate queries.
Recently, with the rise of natural language generation, some works cast KBQA to a Seq2Seq task.
CBR-KBQA \cite{Das2021case} uses T5 to directly transform a question into a SPARQL and outperforms previous \sota~ result by a large margin.
RnG-KBQA \cite{ye2021rng} proposes a Rank-and-Generate approach which first adopts a contrastive ranker to rank candidate logic forms, then generates the final logic form in a Seq2Seq manner.
\begin{table}[ht] 
    \centering
    \small
    \begin{tabular}{p{8.0 cm}}
    \toprule
    \textbf{Question: }What films featuring \textbf{Taylor Swift} have netflix\_id numbers above \textbf{70068848} \\
    \textbf{SubQ1: } what films featuring \textbf{swift} \\
    \textbf{SubQ2: } have netflix\_id numbers above \\
    \textbf{Ours: } What films \des~featuring Taylor Swift \des~have netflix\_id numbers above 70068848 \\
    \midrule
    \textbf{Question: }What schools were attended by the characted of focus in the film ``\textbf{William \& Kate}''\\
    \textbf{SubQ1:} what schools were attended by [INQ] \\
    \textbf{SubQ2:} the characted of ``\textbf{characted focus \& the film}''\\
    \textbf{Ours:} What schools \des~were attended by \inql~the characted of focus in the film ``William \& Kate'' \inqr\\
    \bottomrule
    \end{tabular}
    \caption{Failed cases of another generative method HSP \cite{Zhang2019complex}, compared with our results. HSP splits each question into two sub-questions, namely SubQ1 and SubQ2. The bold words indicate the missing tokens or unexpected tokens that are in conflict between the 
    original question and sub-questions. 
    }
    \label{tab:seq2seqqd_error_case}
\end{table}

In this paper, we propose \qdtqa~following the promising SP-based paradigm.
By incorporating question decomposition into KBQA system, we outperform \sota~methods on both CWQ and LC datasets.

\subsection{Question Decomposition}

Question decomposition essentially provides an ungrounded query graph that handles structure disambiguation. 
With the help of the question's structure, the QA system can avoid inefficient traversal of relation paths \cite{Chen2020Formal}.
There are mainly three kinds of question decomposition methods:

(1) Splitting-based methods, such as SplitQA \cite{Talmor2018the} and DecompRC \cite{Min2019multi}, adopt pointer network to split a question into two parts. While these methods can preserve the original questions, they are too specific to support more complex structure.

(2) Generative methods are more flexible and can be easily extended to different target formats. 
HSP \cite{Zhang2019complex} employs a Seq2Seq model with copy mechanism to generate sub-questions. 
However, these methods still decompose a question into two parts and
can not guarantee that the sentence semantics stay unchanged (as shown in Table \ref{tab:seq2seqqd_error_case}).
As a result, they may lose some tokens or generate unexpected tokens, which would corrupt the semantics of the input question 
and pose difficulties in evaluating the performance.

(3) Rule-based methods. A representative work is EDG \cite{hu2021edg}. It iteratively transforms the constituency tree into an entity-centric graph with hand-crafted rules.  
Generally, this method has limited coverage and is heavily dependent on constituency parsing. 
Moreover, the iterative decomposition approach lacks global awareness and can lead to error cascade when further decomposing sub-questions. 
EDG can handle questions with multiple compositionality types, but how to generate the complex structure in a neural way is challenging.

\begin{sloppypar}
To better model the structure of question,
in this paper,
we propose a tree-based decomposition structure called QDT, along with a two-staged method \cluedecipher~to generate QDT.
\cluedecipher~can leverage the advantages of both splitting-based methods (questions remaining unchanged) and generative methods (strong generation ability and flexibility).

\end{sloppypar}

\section{Question Decomposition Tree}
\label{sec:question decomposition tree}
In this section, we propose Question Decomposition Tree (QDT), a tree-based structure, to model the decomposition structure of complex question.
We first define QDT and then present our QDT dataset, named \qdtrees.

\subsection{Definitions}
\label{subsec:qdt_definitions}
Given a question $q$, the QDT of $q$ is a tree containing two types of nodes: the question node and the description node. 
As shown in the example in the lower part of Figure \ref{fig:main_example}, the root question node (green oval) indicates the original question,
whose answer is constrained by two description nodes, \textit{D1} and \textit{D2}~(dashed rectangle).
Note that \textit{D2} contains a placeholder ``[INQ]'', indicating the existence of an inner question. 
The inner question is represented by a subtree,
the root of which is an inner question node (red oval) pointed by \textit{D2}.
Meanwhile, the inner question node points to two description nodes, \textit{D3} and \textit{D4}~(dashed rectangle).
The structure of QDT enables it to represent the combinations of multiple compositionality types.
We also provide an equivalent linear representation of QDT under the tree illustration by introducing three separators, i.e., \inql, \inqr, \des. 
Among them, \inql~and \inqr~indicate the left and right boundaries of an inner question, while \des~splits conjunctive descriptions.

\begin{sloppypar}
A similar work to QDT is Entity Description Graph (EDG) \cite{hu2021edg},
an entity-centric graph structure.
Compared to EDG, QDT has three differences. 
Firstly, QDT supports more kinds of questions. For example, EDG cannot represent the questions whose answers are not entities, e.g., \textit{``Did \inql~Curry join the Warriors \inqr~before \inql~LeBron played for the Lakers \inqr?''}.
Secondly, the structure of EDG (i.e., a graph with three types of nodes and three types of edges) is complicated, rendering it hard to train with a neural model.
In comparison, QDT is more concise and has an equivalent linearized representation (as indicated in Figure \ref{fig:main_example}), making it easier to generate tree-based structure in a neural way.
Thirdly, QDT keeps the original question unchanged, except inserting some tags, while EDG removes some dummy words, such as \textit{which, that}, etc. 
\end{sloppypar}

\subsection{QDT Dataset}
\label{subsec:QDT Dataset}

To train a model for question decomposition and provide a benchmark, we construct a dataset called \qdtrees, with 6,607 Question Decomposition Trees of complex questions from existing KBQA datasets.

\begin{figure*}[htb]
    \centering
    \includegraphics[scale=0.795]{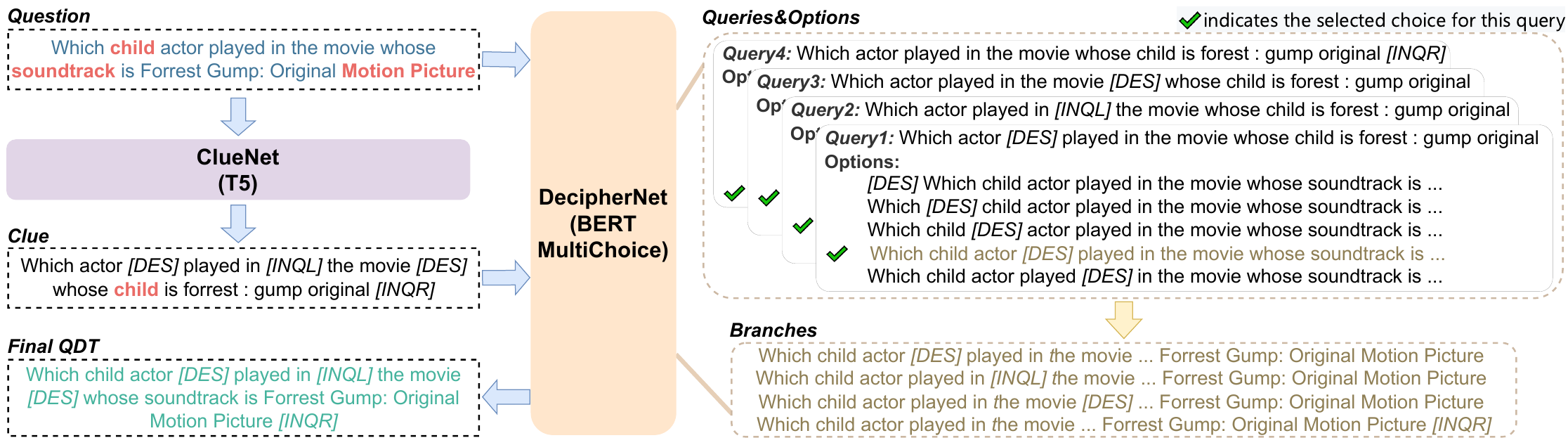}
    \caption{An overview of our two-staged decomposition method \cluedecipher,~consisting of \cluenet~and \deciphernet. The tokens in red bold are the missing or unexpected tokens that are in conflict between question and clue.}
    \label{fig:clue_decipher_process}
\end{figure*}

\subsubsection{Data Collection}
\label{subsec:data collection}
\begin{sloppypar}
The questions in \qdtrees~are derived from two
complex KBQA datasets: ComplexWebQuestions (CWQ) \cite{Talmor2018the} and LC-QuAD 1.0 (LC) \cite{trivedi2017lc}.
For CWQ, we annotate three subsets with 2,000/500/500 questions randomly sampled from the training/validation/test sets, respectively. 
Following \cite{Talmor2018the}, we regard the comparative and superlative questions as conjunction question.
Since LC does not provide an official validation set, we split the training set into a new training set (the first 3,200 questions) and a validation set (the last 800 questions). 
We annotate all complex questions of LC which have more than one triple pattern in SPARQL queries (3,307 questions).
Finally, we obtain 6,607 examples in total, detailed in Table \ref{tab:QDTrees statistics}.

\end{sloppypar}

\begin{table}[t]
    \centering
        \begin{tabular}{l|r|r|r|r} 
        \toprule
            \textbf{Source}  &\textbf{Comp.}  & \textbf{Conj.} &\textbf{Comp.\&Conj.} & \textbf{Total} \\ 
           \hline
           
           CWQ          & 1,350  & 2,864  & 1,214 & 3,000   \\ \hline
           \quad Train            & 900    & 1,916    & 816   & 2,000 \\
           \quad Dev              & 225    & 473    & 198   & 500     \\
           \quad Test             & 225    & 475    & 200   & 500     \\ \midrule
           LC       & 1,958  & 3,002  & 1,554 & 3,607   \\ \hline
           \quad Train             & 1,270  & 1,895  & 983   & 2,313   \\
           \quad Dev               & 308    & 487    & 246   & 576     \\
           \quad Test              & 380    & 620    & 325   & 718     \\  \midrule
           Total             & 3,308  & 6,067  & 2,768 & 6,607   \\ 
        \bottomrule
        \end{tabular}
    \caption{Statistics of \qdtrees.
    Comp. and Conj. refer to the number of questions containing the compositionality type of composition and conjunction, respectively.
    }
    \label{tab:QDTrees statistics}
\end{table}

\subsubsection{Annotation} 
\label{subsec:dataset annotation}
We invite four graduate students, majored in Computer Science and familiar with KBQA, to annotate \qdtrees.
Before annotation, they are informed of the detailed instructions with clear examples. 
The annotation consists of two phases.
In the annotation phase, two annotators independently label linearized QDT for all examples. 
Auxiliary information, including machine questions, intermediary questions, and SPARQL queries, is provided to help understand the meaning of the question.
In the validation phase, for the examples with consistent annotations (exact-match inter-annotator agreement is 0.92) in the annotation phrase, the other two annotators check if they agree with the results.
For the inconsistent examples, all annotators discuss and determine the final annotation.
The annotation lasts for two weeks and takes around 60 hours for each annotator.

\section{Question Decomposition Method}
\label{sec:question_decomposition_method} 
\begin{sloppypar}
To obtain the tree-based structure, 
we propose a two-staged decomposition method called \cluedecipher.
A running example of \cluedecipher~is shown in Figure \ref{fig:clue_decipher_process}.
Instead of deriving the generated decomposition result directly, we regard it as a clue for inserting separators into the original question. 
Our method consists of \cluenet, used to generate a preliminary decomposition result as a clue, and \deciphernet, used to obtain the inserting positions of separators in the original question. 
\end{sloppypar}
 
\subsection{ClueNet}
\label{subsec:clue_net}
\cluenet~aims to generate a \textit{Clue} for a question.
We use T5 \cite{Raffel2020t5} as 
basic model. 
A \textit{Clue} is a preliminary decomposition which is literally a corrupted question with some separators.
In other words, other than the separators, a \textit{Clue} may have some differences from the original question.
As shown in the example in Figure \ref{fig:clue_decipher_process}, the \textit{Clue} loses ``Motion Picture''  and  part of type constrain (``child actor'' is changed to 
 ``actor'').
Besides, it fails to generate the relation (whose \textbf{soundtrack} is) between ``the movie'' and ``Forrest Gump: Original Motion Picture'' in the original question, but instead generates a relation (whose \textbf{child} is) that does not exist.

\subsection{DecipherNet}
\deciphernet~aims to locate the inserting position of each separator in the original question, according to the \textit{Clue} obtained from \cluenet. 
Separators in the \textit{Clue} are processed in turn.
As shown in the example in Figure \ref{fig:clue_decipher_process}, we first break the \textit{Clue} down into several \textit{Queries} by preserving only one separator at a time.
As a result, a \textit{Clue} with $k$ separators derives $k$ \textit{Queries}.

For each \textit{Query}, we construct five \textit{Options} indicating the possible positions for a separator in the original question.
After that, \deciphernet~selects the most possible option, called \textit{Branch}, such process is formulated as a multiple-choice task.
For \textit{Query1} in Figure \ref{fig:clue_decipher_process}, we first estimate that the approximate inserting position of the first \des~
is 2, according to its token position. 
We provide five \textit{Options} by inserting \des~into the original question in positions from 0 to 4. 
Then, \deciphernet~selects the best choice (\textit{Branch}), from the \textit{Options}.
Compared to the \textit{Query}, with a separator in a corrupted question, the \textit{Branch} is presented by inserting a separator into the original question.
After obtaining all the \textit{Branches}, we merge them to form a QDT.

\subsection{Training Data Collection}
\begin{sloppypar}

\deciphernet~ 
selects the \textit{Branch} for each \textit{Query} from some \textit{Options}. 
Note that the only difference between \textit{Query} and \textit{Branch} is that a \textit{Query} is a corrupted question with a separator, while a \textit{Branch} is the original question with a separator.
Therefore, we formulate the training data collection in three steps:
Firstly, we break a golden QDT down into several golden \textit{Branches} by retaining only one separator at a time.
Secondly, we construct some negative \textit{Options} for each \textit{Branch} by shifting the separator to other 4 neighboring positions.
Thirdly, we corrupt the \textit{Branches} to construct the \textit{Queries}. 
For each token in a \textit{Branch}, we have 4 ways to corrupt it, the probability of each way is a predefined hyper-parameter given in parentheses: 
(1)  replace this token with a random word in this \textit{Branch} (1\%), 
(2)  delete this token (1\%), 
(3)  add another random token in this \textit{Branch} after this token (1\%),
(4)  convert this token to its vocabulary id and reconvert the id to the token by the same tokenizer (97\%). 
\end{sloppypar}

\begin{sloppypar}

\section{Evaluation of Decomposition Method}
\label{subsec:decomposition_evaluation}

\begin{sloppypar}
We evaluate the decomposition quality of different decomposition methods from two aspects: tree-based evaluation and sequence-based evaluation.
\end{sloppypar}

\subsection{Baselines}
We compare our method with four baseline methods. 
(1) SplitQA \cite{Talmor2018the} finds split points with a pointer network.
(2) DecompRC \cite{Min2019multi} formulates decomposition as a span prediction task.
(3) HSP \cite{Zhang2019complex} leverages a Seq2Seq model to generate sub-questions. 
(4) EDGQA \cite{hu2021edg} employs hand-crafted rules to generate a graph to represent the question. 
The first three methods are sequence-based methods which are designed to split a question into two sub-questions while EDGQA is a graph-based method.
\end{sloppypar}

\subsection{Implementation Details}
Our decomposition model is based on Pytorch \cite{adam2019pytorch} and Hugging Face \cite{wolf2020transformers}. We use T5-base with Adafactor optimizer for \cluenet, and BERT-base with SGD optimizer for \deciphernet.
The batch sizes for \cluenet~and \deciphernet~are set to 64.
We train our models for 100 epochs on an NVIDIA GeForce RTX 3090 GPU and save the best checkpoints on the validation set.
All methods except EDGQA are trained and tested on \qdtrees.

\subsection{Tree-based Evaluation}
\label{subsubsec:qdt-based evaluation}

\subsubsection{Metrics}
We consider three metrics: 

\textbf{Exact Match (EM)} denotes whether the predicted decomposition is completely the same as the golden one.

\textbf{Tree Depth Accuracy (TDA)} denotes the ratio of generated decomposition whose depth is equal to the golden ones. 

\textbf{Graph Edit Distance (GED)\footnote{We use the implementation provided by \url{https://networkx.org/}}} 
is to measure the minimal transition cost from the predicted decomposition to the golden one.
The lower the GED score, the better. 
Three edit operations (addition, deletion, and substitution) are considered, with
predefined cost following \cite{Wolfson2020break}. 

\begin{sloppypar}
\subsubsection{Results}
As shown in Table \ref{tab:tree_eval}. Our method achieves 0.8332 on EM, and significantly outperforms EDGQA on GED and TDA. 
We do not report the EM of EDGQA because it cannot be trained on \qdtrees~and removes some dummy words, making it hard to match with our annotation.

To demonstrate the effectiveness of our \deciphernet, we perform an ablation study by removing \deciphernet~from \cluedecipher. 
Results show that \cluedecipher~is superior to the bare \cluenet~by 2.02\% on tree-based EM.
This suggests that our proposed \deciphernet~can further promote generative method. 
Besides, we also find that none of the example become worse after the incorporation of \deciphernet, which means \deciphernet~is stable and safe as an external module for generative decomposition method.
Since we adopt separator-insertion according to \textit{Clue}, all generated QDTs do not change any information in the original question, which prevents downstream tasks from error propagation.
\end{sloppypar}

\begin{table}[!tb]
    \centering
    \begin{tabular}{l|ccc}
    \toprule
     Method & EM  & TDA & GED  \\ \midrule
     EDGQA \cite{hu2021edg} & -  & 0.7315 & 0.3799\\ \midrule
     \cluedecipher &\textbf{0.8332} &\textbf{0.9650} & \textbf{0.0554} \\
     \hspace{1em} w/o \deciphernet   & 0.8130 & \textbf{0.9650} & 0.0558  \\
     \bottomrule
    \end{tabular}%
    \caption{Tree-based Decomposition evaluation.}
    \label{tab:tree_eval}
\end{table}
\begin{table}[!tb]
    \centering
    \setlength{\tabcolsep}{0.5mm}{
        \begin{tabular}{l|ccc}
            \toprule
            Method    & EM    & BLEU & ROUGE   \\  \midrule
            SplitQA \cite{Talmor2018the}   & 0.653 & 0.734  & 0.905   \\
            DecompRC \cite{Min2019multi}  & 0.862 & 0.954  & 0.988  \\
            HSP \cite{Zhang2019complex}      & 0.252 & 0.679  & 0.881   \\ 
            HSP\,+\,\deciphernet                 & 0.793  & 0.935 & 0.983\\
            \midrule             
            \cluedecipher & \textbf{0.909}   & \textbf{0.970}  & \textbf{0.993}  \\
            \hspace{1em} w/o \deciphernet & 0.889 & 0.966 & 0.991  \\
             \bottomrule
        \end{tabular}%
     }
    \caption{Sequence-based Decomposition evaluation.}
    \label{tab:seq_eval}
\end{table}

\subsection{Sequence-based Evaluation}
\label{subsubsec:subquestion-based evaluation}
\subsubsection{Metrics} To compare with sequence-based methods, we degrade \cluedecipher~to decompose a question into only two sub-questions. 
We evaluate the performance 
from two aspects: Exact Match and Text Similarities.
In order to make a comprehensive evaluation with other generative methods, we use two text similarity metrics introduced in HSP \cite{Zhang2019complex}, i.e., BLEU-4 \cite{papineni2002bleu} and ROUGE-L \cite{lin2004rouge}.

\subsubsection{Results}
Table \ref{tab:seq_eval} shows that even with the degradation of generating only 2 parts, \cluedecipher~still consistently surpasses other methods on both Exact Match and Text Similarities. 
Concretely, our method achieves the highest EM score of 0.909, exceeding DecompRC by 4.7\%.
Besides, by removing \deciphernet, the sequence-based EM decreases from 0.909 to 0.889.
Among the baselines, HSP is a generative-based decomposition method and achieves a poor performance on EM.
Intuitively, our \deciphernet~can be leveraged to revise its result.
After using \deciphernet, the performance increases notably on all three metrics. 
It shows that our \deciphernet~is easy to be adapted to other generative methods and reveal their potential performance.

\begin{figure*}[htb]
    \centering
    \includegraphics[scale=1.5]{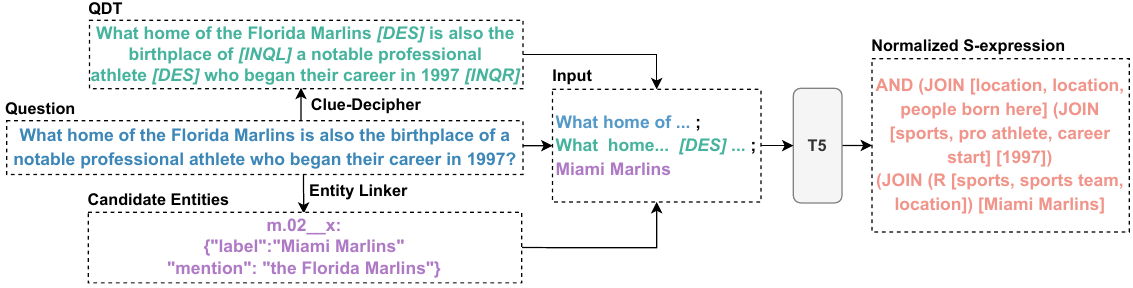}
    \caption{An overview of \qdtqa. 
    }
    \label{fig:s2s_process}
\end{figure*}

\section{KBQA based on QDT} 

In this section, we present \qdtqa, following the Seq2Seq manner and employing QDT to promote the performance.
The framework is shown in Figure \ref{fig:s2s_process}. 
\qdtqa~receives the concatenation of question, QDT, and candidate entities as input and yields an executable query. 

\subsection{Entity Linking}  
To get candidate entities, we adopt off-the-shelf tools: ELQ \cite{li2020efficient} and FACC1 \cite{gabrilovich2013facc1}. 
ELQ is an end-to-end linking model by dense retrieval and returns candidate entities $CandEnt_{ELQ}$ with confidence scores.  
FACC1 is a large mapping from entity mention to Freebase ID. 
We utilize a NER model provided by GrailQA \cite{gu2021beyond} to detect mentions and match with FACC1 through string similarity to get $CandEnt_{FACC1}$. 
We adopt a BERT-based sentence classification model to rank $CandEnt_{FACC1}$ and merge the result with $CandEnt_{ELQ}$ to get the final candidate entities.

\subsection{Query Normalization}
\begin{sloppypar}
In order to better adapt Pre-trained Language Model (PLM) to a KBQA task, 
we consider \textbf{normalized S-expression} as target format of query generation model instead of SPARQL.
The conversion from SPARQL to S-expression is accomplished via the 
script provided by \cite{ye2021rng}.
After that, we make some modifications to the entities and relations of S-expression to get the normalized S-expression.
In detail, for each entity, the Freebase ID is mapped to its label via \textit{rdfs:labels}. 
For each relation, it is converted into a word list which is closer to nature language.
Furthermore, we surround each KB element, including entity, relation, and literal, 
with a pair of brackets to remind the language model that this is a KB element.
For example, in Figure \ref{fig:s2s_process}, the Freebase entity ``m.02\_\_x'' is mapped to  ``[Miami Marlins]'' and the relation ``location.location.people\_born\_here''
is converted to `` [location, location, people born here]''.
\end{sloppypar}

\subsection{Training Query Generation Model} 
Following \cite{Das2021case},
we employ T5-base as our backbone query generation model. Given the question $Q$ and its golden SPARQL $sq$,
we first convert $sq$ to normalized S-expression $Sexp$, according to Query Normalization.
After that, we extract golden entities $Ent$ from $sq$, then map each entity $e$ in $Ent$ to its label $e_l$ to obtain $Ent_{label}$.
We construct the model input by concatenating $Q$ with the linearized QDT $QDT$ and $Ent_{label}$. 
Such input is fed to T5 encoder to obtain the representation $H_{Sexp'}$. After that, the T5 decoder decodes $H_{Sexp'}$ to a logic form $Sexp'$ token by token. 
We optimize the cross-entropy loss between $Sexp'$ and $Sexp$ using teacher forcing.
\begin{equation}
    H_{Sexp'} = T5Encoder([Q;QDT;Ent_{label}])
\end{equation}
\begin{equation}
    Sexp' = T5Decoder(H_{Sexp'})
\end{equation}

\subsection{Inference}
During inference, we first conduct entity linking to get candidate entities and concatenate them with question and QDT as the input of the query generation model. 
After obtaining the normalized S-expression, we convert each normalized element, including the entity label, modified relation, and literal, to its KB representation to get S-expression. 
The S-expression is converted to an executable SPARQL query by the script provided by \cite{gu2021beyond}.
The SPARQLs that can not be executed against KB are skipped. 
We regard the result of the first SPARQL that returns a non-empty result as the final answer.

\section{Evaluation of KBQA Method}
In this section, we evaluate the performance of \qdtqa~and apply \cluedecipher~to another KBQA system: EDGQA \cite{hu2021edg}.

\subsection{Implementation Details}

\subsubsection{\qdtqa}: We employ T5-base as the query generation model and train with AdamW optimizer. The batch size is set to 16 and the max length is set to 196. The entity disambiguation model is 
based on BERT-base, in which we set batch size to 16 and max length to 96. 

\subsubsection{EDGQA\,+\,\cluedecipher}: We replace the decomposition method in EDGQA with \cluedecipher.
Other settings remain the same  for fair comparison.

\subsection{Datasets}
We conduct experiments on two popular complex question answering datasets: CWQ (for \qdtqa) and LC (for EDGQA\,+\,\cluedecipher). 
\subsubsection{ComlexWebQuestions (CWQ)}\cite{Talmor2018the} contains 34,689 complex questions over Freebase (version 2015-08-09). The train/validation/test sets contain 27,639/3,519/3,531 questions, respectively. 
\subsubsection{LC-QuAD 1.0 (LC)} \cite{trivedi2017lc} contains 5,000 questions over DBpedia (2016-04), with 4,000 training and 1,000 testing questions.
72.64\% questions of LC are complex questions,
which means the corresponding SPARQL queries have more than one triple patterns. 

\subsection{Baselines}

We compare \qdtqa~with the following three systems on CWQ.
(1) \cite{qin2021improving} gradually shrinks knowledge base to a desired query graph.
(2) \cite{huang2021unseen} directly generate SPARQL query without simplification.
(3) CBR-KBQA \cite{Das2021case} generates logical forms by retrieving relevant cases.

EDGQA\,+\,\cluedecipher~is compared with:
(1) EDGQA \cite{hu2021edg} generates sub-queries node-by-node following a decomposition, and then composes the final query.
(2) NSQA \cite{Kapanipathi2021leveraging} adopts AMR to get shallow semantic parsing of questions.  
(3) \cite{liang21bert} propose a BERT-based decoder to extract triple patterns. 
(4) STaG-QA \cite{Ravishankar2022atwo} separates the semantic parsing process from knowledge base interaction and achieves notable performance on multiple KBs.

QA results of baselines are from their original paper.

\subsection{Metrics}
We follow the metrics considered in baseline methods.
For CWQ, we use Average F1 (Avg. F1) and Accuracy (Acc).
For LC, we use Precision (P), Recall (R) and Macro F1 (F1).

\subsection{Result and Analysis on \qdtqa}
\label{subsec:s2s_evaluation}

\subsubsection{Main Results} 
\begin{sloppypar}

The result of \qdtqa~on CWQ is summerized in Table \ref{tab:seq2seq_qa}. \qdtqa~significantly outperforms previous methods. 
To verify the effectiveness of question decomposition in \qdtqa, we remove QDT from the input.
Result shows that, without QDT, \qdtqa~drops 1.3\% on average F1, which illustrates that our decomposition can improve the performance of downstream KBQA task.
We also evaluate the performance of a model, replacing tree-based decomposition with sequence-based decomposition, with an average F1 decrease of 0.8\%.
It demonstrates that tree-based decomposition can provide richer information and is important for the representation of complex questions.

\subsubsection{The Effectiveness of Question Decomposition} 
We further analyze how QDT contributes to the result, from two critical factors of KBQA, i.e., structure and relation linking. 
We analyze all questions on the test set of CWQ.
After using QDT, 
additional 7.31\% questions derive correct query structures. 
We believe that QDT can provide a structural guidance to avoid invalid searching of noisy query structures.
Besides, relation linking can also benefit from question decomposition.
Since a decomposition, beforehand, separates different semantic representations,
which reduces the search space and avoids the distraction of other relations. 
In fact, 10.53\% questions get better relation linking results after the incorporation of QDT.
\end{sloppypar}

\subsubsection{The Impact of Different Decomposition Methods}
\begin{sloppypar}
To evaluate the impact of different decomposition methods on QA system,
we replace the decomposition result in \qdtqa~with four other decomposition methods. Results in Table \ref{tab:seq2seq_qa} demonstrate that our decomposition consistently outperforms other methods. 
Combining the results of the previous experiments, it shows that our method is superior to others on both decomposition and question answering.
\end{sloppypar}

 \begin{table}[!tb] 
    \centering    
    \begin{tabular}{l|cc}
    \toprule
     Method & Avg. F1 & Acc  \\  \midrule
    \cite{qin2021improving} &0.462 & -\\
     \cite{huang2021unseen} & 0.682 &  - \\
     T5-11B + Revise \cite{Das2021case} & 0.582 & 0.556 \\
     CBR-KBQA \cite{Das2021case}& 0.700 & 0.671 \\ \midrule
     \qdtqa &  \textbf{0.728} & \textbf{0.679} \\  
      \hspace{1em} w/o QDT & 0.715 & 0.666\\ 
      \hspace{1em} w/o tree-based structure & 0.720 & 0.670\\
      
    \midrule
     \hspace{1em} w/ SplitQA  & 0.716 &  0.669 \\
     \hspace{1em} w/ DecompoRC  & 0.716 & 0.669 \\ 
     \hspace{1em} w/ HSP  & 0.717 & 0.669 \\
     \hspace{1em} w/ EDGQA  & 0.714 & 0.665 \\
     \bottomrule
    \end{tabular}%
    \caption{QA performance of \qdtqa~ on CWQ compared with baselines. We also report the performance with different decomposition methods. 
    }
    \label{tab:seq2seq_qa}
\end{table}

\subsubsection{Error Analysis}
\label{subsubsec:s2s_error_analyze}
We randomly sample 100 questions that are incorrectly answered by \qdtqa~(F1 \textless 1.0) and categorize them into three types.
 
\textbf{Question Decomposition (5\%)}. We manually verify whether a decomposition is acceptable under the same standard for annotation.  
For example, \textit{``What movies have been written by \inql~authors of \des~monty python''} is a failed case.

\textbf{Query Structure (44\%)}.
Even with a correct decomposition structure, it is still possible for \qdtqa~to generate an S-expression that does not match the golden one at the structure level. 
The reason lies on the fact that \qdtqa~utilizes QDT as an implicit restriction rather than forcing the system to strictly follow this structure.  

\textbf{Entity and Relation Linking (51\%)}.
Linking is still the main obstacle for KBQA. There are 26\% and 25\% errors caused by entity and relation linking, respectively.

\subsection{Result on EDGQA\,+\,\cluedecipher} 
To further verify that our decomposition method can enhance existing decomposition-based KBQA system, we conduct another experiment, incorporating our decomposition result on a \sota~decomposition-based KBQA system, i.e., EDGQA \cite{hu2021edg}.  
We adopt EDGQA  as a basic system for two reasons. Firstly, it is the \sota~decomposition-based system on a complex question answering dataset (LC). Secondly, it is the only open-source decomposition-based KBQA system in recent years.

Table \ref{tab:edgqa_different_decomposition} compares EDGQA\,+\,\cluedecipher~with other methods on LC.
Results show that our decomposition significantly improves EDGQA by 11\% on F1 and sets a new \sota~on LC, which demonstrates the utility of our decomposition method.
We also report the performance of different decomposition methods on EDGQA in Table \ref{tab:edgqa_different_decomposition}.
The baseline methods are the same as discussed in decomposition evaluation. 
\cluedecipher~consistently outperforms other decomposition methods, which further verifies the superiority of our decomposition method.

\begin{table}[!tb] 
    \centering 
    \setlength{\tabcolsep}{1.3mm}{
        \begin{tabular}{l|cccc}
        \toprule
         Method & P & R & F1 & $\Delta \rm{F1}$ \\  \midrule
        NSQA   &0.448  & 0.458  & 0.445 & -  \\ 
        STaG-QA & \textbf{0.745*} & 0.548 & 0.536&-\\
        \cite{liang21bert}  & 0.511 & 0.593 & 0.549 &- \\  \midrule
        EDGQA  & 0.505 & 0.560 & 0.531 & 0\\ 
         \hspace{1em} w/ SplitQA & 0.496 & 0.576 & 0.533 & +\,0.002 \\
         \hspace{1em} w/ DecompRC & 0.521 & 0.609 & 0.561 & +\,0.030\\ 
         \hspace{1em} w/ HSP  & 0.433	& 0.507 & 0.467 & -\,0.064 \\
         \hspace{1em} w/ \cluedecipher  & 0.548 & \textbf{0.635} &\textbf{0.588} &\textbf{+\,0.056}\\
         \bottomrule
        \end{tabular} 
    } 
    \caption{ 
    QA performance of baseline methods and EDGQA\,+\,\cluedecipher~ on LC. We also replace \cluedecipher~with other decomposition methods. 
    * indicates that when calculating P, STaG-QA defines the empty answer to have P=1, different from others.
    }
    \label{tab:edgqa_different_decomposition}
\end{table}
\section{Conclusion}
\label{sec:conclusion}
\begin{sloppypar} 
In this paper, we focus on how to make question decomposition an effective way to answer complex questions over KBs.  
To summarize, 
we make the following contributions:
\begin{itemize} 
    \item We propose Question Decomposition Tree (QDT) to represent the decomposition structure of complex questions, along with a dataset \qdtrees~with 6,607 QDTs. We also present a linearized representation of QDT, making it easier to be generated in a neural way.  
    \item We propose \cluedecipher~to generate QDTs that consistently surpass other methods from multiple aspects.  
    It leverages the strong generation ability of NLG models and ensures that the original question remains unchanged. 
    \item We design a QDT-based KBQA system called \qdtqa~which achieves \sota~performance on CWQ.  
    Besides, we demonstrate that our decomposition can further enhance an existing system on LC by 11\%. 
\end{itemize}

\end{sloppypar}

\begin{sloppypar} 
In the future, an interesting topic is to automatically generate decomposition annotations from 
 existing pairs of questions and SPARQLs.
Besides, QDT can be extended to model other linguistic phenomena, such as disjunction, which is worth exploring. 
\end{sloppypar}

\section*{Acknowledgements}

This work was supported by the National Natural Science Foundation of China (NSFC) under Grant No. 62072224. We would like to thank Xixin Hu and Xuan Wu for their efforts in the early stages of this work. We also thank the anonymous reviewers for their constructive comments.

\bibliography{aaai23}

\begin{thebibliography}{28}
\providecommand{\natexlab}[1]{#1}

\bibitem[{Auer et~al.(2007)Auer, Bizer, Kobilarov, Lehmann, Cyganiak, and
  Ives}]{auer2007dbpedia}
Auer, S.; Bizer, C.; Kobilarov, G.; Lehmann, J.; Cyganiak, R.; and Ives, Z.
  2007.
\newblock Dbpedia: A nucleus for a web of open data.
\newblock In \emph{The semantic web}, 722--735. Springer.

\bibitem[{Bollacker et~al.(2008)Bollacker, Evans, Paritosh, Sturge, and
  Taylor}]{bollacker2008freebase}
Bollacker, K.; Evans, C.; Paritosh, P.; Sturge, T.; and Taylor, J. 2008.
\newblock Freebase: a collaboratively created graph database for structuring
  human knowledge.
\newblock In \emph{Proceedings of the SIGMOD}, 1247--1250.

\bibitem[{Chen et~al.(2020)Chen, Li, Hua, and Qi}]{Chen2020Formal}
Chen, Y.; Li, H.; Hua, Y.; and Qi, G. 2020.
\newblock Formal Query Building with Query Structure Prediction for Complex
  Question Answering over Knowledge Base.
\newblock In Bessiere, C., ed., \emph{Proceedings of the AAAI}, 3751--3758.
  ijcai.org.

\bibitem[{Das et~al.(2021)Das, Zaheer, Thai, Godbole, Perez, Lee, Tan,
  Polymenakos, and McCallum}]{Das2021case}
Das, R.; Zaheer, M.; Thai, D.; Godbole, A.; Perez, E.; Lee, J.-Y.; Tan, L.;
  Polymenakos, L.; and McCallum, A. 2021.
\newblock {Case-based Reasoning for Natural Language Queries over Knowledge
  Bases}.
\newblock \emph{EMNLP}, 9594--9611.

\bibitem[{Gabrilovich and Subramanya(2013)}]{gabrilovich2013facc1}
Gabrilovich, R.~M., E.; and Subramanya, A. 2013.
\newblock {Facc1: Freebase annotation of clueweb corpora, version 1 (release
  date 2013-06-26, format version 1, correction level 0)}.
\newblock \url{http://lemurproject.org/clueweb09/FACC1/}.
\newblock Accessed: 2013-06-26.

\bibitem[{Gu et~al.(2021)Gu, Kase, Vanni, Sadler, Liang, Yan, and
  Su}]{gu2021beyond}
Gu, Y.; Kase, S.; Vanni, M.; Sadler, B.; Liang, P.; Yan, X.; and Su, Y. 2021.
\newblock Beyond IID: three levels of generalization for question answering on
  knowledge bases.
\newblock In \emph{Proceedings of the Web Conference}, 3477--3488.

\bibitem[{Hu et~al.(2021)Hu, Shu, Huang, and Qu}]{hu2021edg}
Hu, X.; Shu, Y.; Huang, X.; and Qu, Y. 2021.
\newblock EDG-Based Question Decomposition for Complex Question Answering over
  Knowledge Bases.
\newblock In \emph{Proceedings of the ISWC}, 128--145.

\bibitem[{Huang, Kim, and Zou(2021)}]{huang2021unseen}
Huang, X.; Kim, J.-J.; and Zou, B. 2021.
\newblock Unseen Entity Handling in Complex Question Answering over Knowledge
  Base via Language Generation.
\newblock In \emph{Findings of the Association for Computational Linguistics:
  EMNLP 2021}, 547--557. Punta Cana, Dominican Republic: Association for
  Computational Linguistics.

\bibitem[{Kapanipathi et~al.(2021)Kapanipathi, Abdelaziz, Ravishankar, Roukos,
  Gray, Astudillo, Chang, Cornelio, Dana, Fokoue, Garg, Gliozzo, Gurajada,
  Karanam, Khan, Khandelwal, Lee, Li, Luus, Makondo, Mihindukulasooriya,
  Naseem, Neelam, Popa, Reddy, Riegel, Rossiello, Sharma, Bhargav, and
  Yu}]{Kapanipathi2021leveraging}
Kapanipathi, P.; Abdelaziz, I.; Ravishankar, S.; Roukos, S.; Gray, A.;
  Astudillo, R.; Chang, M.; Cornelio, C.; Dana, S.; Fokoue, A.; Garg, D.;
  Gliozzo, A.; Gurajada, S.; Karanam, H.; Khan, N.; Khandelwal, D.; Lee, Y.~S.;
  Li, Y.; Luus, F.; Makondo, N.; Mihindukulasooriya, N.; Naseem, T.; Neelam,
  S.; Popa, L.; Reddy, R.; Riegel, R.; Rossiello, G.; Sharma, U.; Bhargav,
  G.~P.; and Yu, M. 2021.
\newblock {Leveraging Abstract Meaning Representation for Knowledge Base
  Question Answering}.
\newblock \emph{Findings of the Association for Computational Linguistics:
  ACL-IJCNLP 2021}, (c): 3884--3894.

\bibitem[{Lan et~al.(2021)Lan, He, Jiang, Jiang, Zhao, and Wen}]{lan2021survey}
Lan, Y.; He, G.; Jiang, J.; Jiang, J.; Zhao, W.~X.; and Wen, J. 2021.
\newblock A Survey on Complex Knowledge Base Question Answering: Methods,
  Challenges and Solutions.
\newblock In Zhou, Z., ed., \emph{Proceedings of the IJCAI}, 4483--4491.

\bibitem[{Lan and Jiang(2020)}]{lan2020query}
Lan, Y.; and Jiang, J. 2020.
\newblock Query Graph Generation for Answering Multi-hop Complex Questions from
  Knowledge Bases.
\newblock In \emph{Proceedings of the ACL}, 969--974.

\bibitem[{Li et~al.(2020)Li, Min, Iyer, Mehdad, and Yih}]{li2020efficient}
Li, B.~Z.; Min, S.; Iyer, S.; Mehdad, Y.; and Yih, W.-t. 2020.
\newblock Efficient One-Pass End-to-End Entity Linking for Questions.
\newblock In \emph{Proceedings of the EMNLP}, 6433--6441.

\bibitem[{Liang et~al.(2021)Liang, Peng, Yang, Zhao, Liu, and
  McGuinness}]{liang21bert}
Liang, Z.; Peng, Z.; Yang, X.; Zhao, F.; Liu, Y.; and McGuinness, D.~L. 2021.
\newblock BERT-based Semantic Query Graph Extraction for Knowledge Graph
  Question Answering.
\newblock In \emph{Proceedings of the ISWC}.

\bibitem[{Lin(2004)}]{lin2004rouge}
Lin, C.-Y. 2004.
\newblock {ROUGE}: A Package for Automatic Evaluation of Summaries.
\newblock In \emph{Text Summarization Branches Out}, 74--81. Barcelona, Spain:
  Association for Computational Linguistics.

\bibitem[{Min et~al.(2019)Min, Zhong, Zettlemoyer, and
  Hajishirzi}]{Min2019multi}
Min, S.; Zhong, V.; Zettlemoyer, L.; and Hajishirzi, H. 2019.
\newblock {Multi-hop reading comprehension through question decomposition and
  rescoring}.
\newblock In \emph{Proceedings of the ACL}, 6097--6109.

\bibitem[{Papineni et~al.(2002)Papineni, Roukos, Ward, and
  Zhu}]{papineni2002bleu}
Papineni, K.; Roukos, S.; Ward, T.; and Zhu, W.-J. 2002.
\newblock {B}leu: a Method for Automatic Evaluation of Machine Translation.
\newblock In \emph{Proceedings of the 40th Annual Meeting of the Association
  for Computational Linguistics}, 311--318. Philadelphia, Pennsylvania, USA:
  Association for Computational Linguistics.

\bibitem[{Paszke et~al.(2019)Paszke, Gross, Massa, Lerer, Bradbury, Chanan,
  Killeen, Lin, Gimelshein, Antiga, Desmaison, Kopf, Yang, DeVito, Raison,
  Tejani, Chilamkurthy, Steiner, Fang, Bai, and Chintala}]{adam2019pytorch}
Paszke, A.; Gross, S.; Massa, F.; Lerer, A.; Bradbury, J.; Chanan, G.; Killeen,
  T.; Lin, Z.; Gimelshein, N.; Antiga, L.; Desmaison, A.; Kopf, A.; Yang, E.;
  DeVito, Z.; Raison, M.; Tejani, A.; Chilamkurthy, S.; Steiner, B.; Fang, L.;
  Bai, J.; and Chintala, S. 2019.
\newblock PyTorch: An Imperative Style, High-Performance Deep Learning Library.
\newblock In Wallach, H.; Larochelle, H.; Beygelzimer, A.; d\textquotesingle
  Alch\'{e}-Buc, F.; Fox, E.; and Garnett, R., eds., \emph{Advances in Neural
  Information Processing Systems}, volume~32. Curran Associates, Inc.

\bibitem[{Qin et~al.(2021)Qin, Li, Pavlu, and Aslam}]{qin2021improving}
Qin, K.; Li, C.; Pavlu, V.; and Aslam, J.~A. 2021.
\newblock {Improving Query Graph Generation for Complex Question Answering over
  Knowledge Base}.
\newblock \emph{EMNLP}, 4201--4207.

\bibitem[{Raffel et~al.(2020)Raffel, Shazeer, Roberts, Lee, Narang, Matena,
  Zhou, Li, and Liu}]{Raffel2020t5}
Raffel, C.; Shazeer, N.; Roberts, A.; Lee, K.; Narang, S.; Matena, M.; Zhou,
  Y.; Li, W.; and Liu, P.~J. 2020.
\newblock {T5: Exploring the limits of transfer learning with a unified
  text-to-text transformer}.
\newblock \emph{Journal of Machine Learning Research}, 21: 1--67.

\bibitem[{Ravishankar et~al.(2022)Ravishankar, Thai, Abdelaziz,
  Mihidukulasooriya, Naseem, Kapanipathi, Rossilleo, and
  Fokoue}]{Ravishankar2022atwo}
Ravishankar, S.; Thai, J.; Abdelaziz, I.; Mihidukulasooriya, N.; Naseem, T.;
  Kapanipathi, P.; Rossilleo, G.; and Fokoue, A. 2022.
\newblock {A Two-Stage Approach towards Generalization in Knowledge Base
  Question Answering}.
\newblock \emph{AAAI}.

\bibitem[{Sun, Bedrax-Weiss, and Cohen(2019)}]{sun2019pullnet}
Sun, H.; Bedrax-Weiss, T.; and Cohen, W. 2019.
\newblock PullNet: Open Domain Question Answering with Iterative Retrieval on
  Knowledge Bases and Text.
\newblock In \emph{Proceedings of the EMNLP-IJCNLP}, 2380--2390.

\bibitem[{Talmor and Berant(2018)}]{Talmor2018the}
Talmor, A.; and Berant, J. 2018.
\newblock {The web as a knowledge-base for answering complex questions}.
\newblock In \emph{Proceedings of the NAACL-HLT}, 641--651.

\bibitem[{Trivedi et~al.(2017)Trivedi, Maheshwari, Dubey, and
  Lehmann}]{trivedi2017lc}
Trivedi, P.; Maheshwari, G.; Dubey, M.; and Lehmann, J. 2017.
\newblock Lc-quad: A corpus for complex question answering over knowledge
  graphs.
\newblock In \emph{the Proceedings of ISWC}, 210--218.

\bibitem[{Wolf et~al.(2020)Wolf, Debut, Sanh, Chaumond, Delangue, Moi, Cistac,
  Rault, Louf, Funtowicz, Davison, Shleifer, von Platen, Ma, Jernite, Plu, Xu,
  Le~Scao, Gugger, Drame, Lhoest, and Rush}]{wolf2020transformers}
Wolf, T.; Debut, L.; Sanh, V.; Chaumond, J.; Delangue, C.; Moi, A.; Cistac, P.;
  Rault, T.; Louf, R.; Funtowicz, M.; Davison, J.; Shleifer, S.; von Platen,
  P.; Ma, C.; Jernite, Y.; Plu, J.; Xu, C.; Le~Scao, T.; Gugger, S.; Drame, M.;
  Lhoest, Q.; and Rush, A. 2020.
\newblock Transformers: State-of-the-Art Natural Language Processing.
\newblock In \emph{Proceedings of the EMNLP: System Demonstrations}, 38--45.
  Online: Association for Computational Linguistics.

\bibitem[{Wolfson et~al.(2020)Wolfson, Geva, Gupta, Goldberg, Gardner, Deutch,
  and Berant}]{Wolfson2020break}
Wolfson, T.; Geva, M.; Gupta, A.; Goldberg, Y.; Gardner, M.; Deutch, D.; and
  Berant, J. 2020.
\newblock Break It Down: {A} Question Understanding Benchmark.
\newblock \emph{Trans. Assoc. Comput. Linguistics}, 8: 183--198.

\bibitem[{Ye et~al.(2022)Ye, Yavuz, Hashimoto, Zhou, and Xiong}]{ye2021rng}
Ye, X.; Yavuz, S.; Hashimoto, K.; Zhou, Y.; and Xiong, C. 2022.
\newblock {RNG}-{KBQA}: Generation Augmented Iterative Ranking for Knowledge
  Base Question Answering.
\newblock In \emph{Proceedings of the 60th Annual Meeting of the Association
  for Computational Linguistics (Volume 1: Long Papers)}, 6032--6043. Dublin,
  Ireland: Association for Computational Linguistics.

\bibitem[{Yih et~al.(2015)Yih, Chang, He, and Gao}]{Yih2015Semantic}
Yih, W.~T.; Chang, M.~W.; He, X.; and Gao, J. 2015.
\newblock {Semantic parsing via staged query graph generation: Question
  answering with knowledge base}.
\newblock In \emph{Proceedings of the ACL-IJCNLP}, 1321--1331.
\newblock ISBN 9781941643723.

\bibitem[{Zhang et~al.(2019)Zhang, Cai, Xu, and Wang}]{Zhang2019complex}
Zhang, H.; Cai, J.; Xu, J.; and Wang, J. 2019.
\newblock {Complex question decomposition for semantic parsing}.
\newblock In \emph{Proceedings of the ACL}, 4477--4486.

\end{thebibliography}
\end{document}